\documentclass{article}
\usepackage[preprint]{colm2026_conference}

\usepackage[utf8]{inputenc}
\usepackage[T1]{fontenc}
\usepackage{amsmath,amssymb,amsthm,mathtools,bm}
\usepackage{microtype}
\usepackage{graphicx,booktabs,placeins}
\usepackage{subcaption}
\usepackage[table]{xcolor}
\usepackage{float}
\usepackage{wrapfig}
\usepackage{listings}
\usepackage[hyphens]{url}
\usepackage{hyperref}
\usepackage[capitalize,noabbrev]{cleveref}
\usepackage{tikz}
\usetikzlibrary{arrows.meta}
\usepackage{arxiv_style}
\usepackage{adjustbox}

\definecolor{darkblue}{HTML}{254E7A}
\definecolor{linkpink}{HTML}{D6336C}
\colorlet{eproblue}{darkblue!8!white}
\hypersetup{colorlinks=true,linkcolor=darkblue,urlcolor=linkpink,citecolor=darkblue}

\lstdefinestyle{eprocode}{
    language=Python,
    basicstyle=\ttfamily\footnotesize,
    keywordstyle=\color{blue!55!black},
    commentstyle=\color{gray!75!black},
    stringstyle=\color{teal!65!black},
    backgroundcolor=\color{gray!3},
    frame=single,
    rulecolor=\color{gray!25},
    framesep=6pt,
    breaklines=true,
    breakatwhitespace=true,
    columns=fullflexible,
    keepspaces=true,
    showstringspaces=false,
    tabsize=4,
    numbers=none,
    captionpos=b,
    aboveskip=10pt,
    belowskip=8pt
}

\newtheorem{theorem}{Theorem}

\newcommand{\E}{\mathbb{E}}

\newcommand{\sg}{\operatorname{sg}}

\title{EPRO: Robust Policy Optimization Through an\\Imperfect Execution-Policy Predictor}
\author{
    \textbf{Yibo Wang}\textsuperscript{1} \thanks{
        Email: \texttt{yibowang2030@u.northwestern.edu}
    } \quad
    \textbf{Changyi Yang}\textsuperscript{2} \quad
    \textbf{Yuanxin Liu}\textsuperscript{1} \quad
    \textbf{Zhaoran Wang}\textsuperscript{1} \thanks{
        Email: \texttt{zhaoranwang@gmail.com}
    }
    \\[0.8em]
    \small
    \textsuperscript{1}Northwestern University \quad
    \textsuperscript{2}Carnegie Mellon University 
}
\begin{document}
\maketitle
\lhead{}
\renewcommand{\headrulewidth}{0pt}
\begin{abstract}
RL for LLMs often separates rollout generation from policy training. Although the two numerical engines share model parameters, they can induce different policies. This creates a challenge beyond correcting a known training–inference mismatch: during learner updates, the execution policy corresponding to the current parameters cannot be directly observed. We view this as optimizing an unobserved execution policy through an imperfect predictor of its evolution. This perspective raises two fundamental questions: \emph{how can we estimate the execution policy's response to learner updates, and how much should we trust that estimate when optimizing the actual execution policy?} We introduce Execution-Policy Robust Optimization (EPRO) to address these challenges. Starting from the observed rollout-time execution policy, we use training-policy changes to construct a differentiable predictor of the current execution policy, whose score yields Transported Score Centering (TSC). Our robust-improvement analysis reveals how prediction error limits the magnitude of reliably justified updates. For practical implementation, we introduce Policy Shift as an observable proxy for prediction risk and use Shift-Aware Soft Clipping (SSC) to attenuate risky updates. Across four Qwen3 model settings, EPRO achieves the highest average held-out accuracy among the compared methods on seven mathematical reasoning benchmarks. Code is available at \url{https://github.com/yibowangnwu/EPRO}.

\end{abstract}








\section{Introduction}
\label{sec:intro}

Reinforcement learning (RL) for large language models commonly separates rollout generation from policy optimization. Inference engines such as vLLM~\citep{kwon2023efficientVLLM} and SGLang~\citep{zheng2024sglang} generate responses, while training frameworks based on FSDP~\citep{zhao2023pytorchFSDP} or Megatron~\citep{shoeybi2019megatron} compute gradients and update model weights. This separation introduces two sources of off-policy discrepancy: \textbf{(i) Training--inference mismatch.} Despite sharing the same model parameters, the two engines can induce different next-token distributions due to differences in numerical kernels, implementations, and precision, resulting in training--inference mismatch (TIM). We therefore distinguish the training policy, used for optimization, from the execution policy, which determines the responses actually generated. \textbf{(ii) Policy lag.} When multiple minibatch updates are performed per rollout, or RL training is asynchronous, the sampling engine may remain at an earlier checkpoint while the training policy continues to update its parameters, introducing policy lag. Existing approaches, including PPO and GRPO, control policy updates through clipped probability ratios \citep{schulman2017proximal,shao2024deepseekmathGRPO}, while more recent methods employ importance-weight truncation, sample masking, trust-region constraints, or score correction~\citep{qi2026rethinkingDPPO,gao2025softSAPO,chen2025minimaxCISPO,yu2026dapo}. Notably, Score Centering (SC) implicitly optimizes the execution policy by correcting training-side scores using the rollout-time execution distribution~\citep{marek2026score}.

\begin{figure}[t]
    \centering

    \begin{adjustbox}{max width=1.0\linewidth,
                      max totalheight=0.5\textheight,
                      keepaspectratio}
        \begin{minipage}{\linewidth}
            \centering

            \begingroup
\definecolor{figureInk}{HTML}{1F2933}
\definecolor{figureMuted}{HTML}{465B71}
\definecolor{figureBlue}{HTML}{254E7A}
\definecolor{figureBlueEdge}{HTML}{8FB6D9}
\definecolor{figureBlueTint}{HTML}{F0F6FB}
\definecolor{figureAmber}{HTML}{955111}
\definecolor{figureAmberEdge}{HTML}{E2A64B}
\definecolor{figureAmberTint}{HTML}{FDF6EC}
\definecolor{figureGradientInk}{HTML}{315D45}
\definecolor{figureGradientEdge}{HTML}{AEC6AA}
\definecolor{figureGradientTint}{HTML}{EAF3E7}
\definecolor{figureNeutralEdge}{HTML}{C6D3E1}

\begin{tikzpicture}[
    x=1pt, y=1pt, text=figureInk,
    panel heading/.style={anchor=north west, inner sep=0pt, align=left,
        font=\fontsize{8.6}{9.4}\selectfont\bfseries},
    detail/.style={inner sep=0pt, align=left, text=figureMuted,
        font=\fontsize{6.8}{7.7}\selectfont},
    stage label/.style={inner sep=0pt, text=figureInk,
        font=\fontsize{7}{8}\selectfont\bfseries},
    engine label/.style={anchor=west, inner sep=0pt,
        font=\fontsize{7.7}{8.1}\selectfont\bfseries},
    card/.style={rounded corners=3pt, line width=0.55pt},
    state/.style={rounded corners=2pt, minimum height=22pt, minimum width=30pt,
        inner sep=1pt, line width=0.65pt},
    minibatch/.style={rounded corners=1.5pt, fill=white, draw=figureAmberEdge!70,
        minimum width=31pt, minimum height=12pt, inner sep=1pt, text=figureAmber,
        font=\fontsize{5.8}{6.8}\selectfont\bfseries},
    wire/.style={draw=figureInk, line width=0.6pt},
    edge/.style={wire, -{Latex[length=3pt]}},
    trainer edge/.style={edge, draw=figureAmber},
    training state/.style={state, minimum width=23pt, minimum height=17pt,
        draw=figureAmberEdge, fill=white, text=figureAmber},
    formula/.style={inner sep=0pt, font=\fontsize{7.5}{9.6}\selectfont}
]
\node[panel heading,font=\fontsize{7.9}{9}\selectfont\bfseries] at (0,174)
    {Problem: Current execution policy is costly to score};
\node[detail,anchor=north west,text width=188pt] at (0,155)
    {Learner updates change both policies, but the\\current execution policy is not
     rescored each minibatch.};
\draw[card,draw=figureBlueEdge!55,fill=figureBlueTint]
    (0,63) rectangle (189,114);
\draw[card,draw=figureAmberEdge!55,fill=figureAmberTint]
    (0,-2) rectangle (189,60);

\node[stage label,anchor=west] at (4,132) {Rollout checkpoint};
\node[stage label] at (164,132) {Current parameters};
\node[inner sep=0pt] at (23,124) {$\theta_0$};
\node[inner sep=0pt] at (167,124) {$\theta$};

\node[engine label,text=figureBlue] at (4,69)
    {Inference engine\enspace{\mdseries\fontsize{6.5}{7.4}\selectfont(Low precision)}};
\node[engine label,text=figureAmber] at (4,53)
    {Training engine\enspace{\mdseries\fontsize{6.5}{7.4}\selectfont(BF16)}};

\node[state,minimum width=28pt,draw=figureBlueEdge,fill=white,text=figureBlue]
    (execution-rollout) at (23,86) {$\mu_{\theta_0}$};
\node[state,minimum width=38pt,minimum height=41pt,draw=figureMuted!55,
      fill=white,densely dashed] (execution-current) at (168,88) {};
\node[inner sep=0pt,text=figureMuted] at (168,100) {$\mu_\theta$};
\node[detail,font=\fontsize{6.3}{7.2}\selectfont\bfseries] at (168,88.5)
    {Unscored};
\node[detail,align=center,font=\fontsize{6}{6.9}\selectfont] at (168,77.5)
    {Extra sync\\and scoring};
\foreach \layer in {2,1} {
    \draw[rounded corners=2pt,draw=figureNeutralEdge!70,fill=figureNeutralEdge!20]
        (96+\layer,76-\layer) rectangle (141+\layer,96-\layer);
}
\node[state,minimum width=45pt,minimum height=20pt,draw=figureNeutralEdge,
      fill=white,align=center,font=\fontsize{6.8}{7.4}\selectfont\bfseries]
    (rollout-batch) at (118.5,86) {Fixed rollout\\batch};
\draw[edge,draw=figureBlue] (execution-rollout.east) -- (rollout-batch.west);

\node[training state] (training-rollout) at (23,13) {$\pi_{\theta_0}$};
\coordinate (update-one) at (43,13);
\node[training state] (training-one) at (64,13) {$\pi_{\theta_1}$};
\coordinate (update-two) at (85,13);
\node[training state] (training-two) at (107,13) {$\pi_{\theta_2}$};
\coordinate (update-final) at (146,13);
\node[training state,minimum width=24pt] (training-current) at (168,13) {$\pi_\theta$};
\draw[trainer edge] (training-rollout.east) -- (update-one) -- (training-one.west);
\draw[trainer edge] (training-one.east) -- (update-two) -- (training-two.west);
\draw[wire,draw=figureAmber,densely dashed] (training-two.east)
    -- node[above,inner sep=1pt,text=figureAmber] {$\cdots$} (update-final);
\draw[trainer edge] (update-final) -- (training-current.west);

\node[minibatch] (minibatch-one) at (43,34) {Minibatch 1};
\node[minibatch] (minibatch-two) at (85,34) {Minibatch 2};
\node[minibatch] (minibatch-final) at (146,34) {Minibatch M};
\coordinate (split-level) at (0,45);
\coordinate (batch-junction) at (rollout-batch.south |- split-level);
\draw[wire] (rollout-batch.south) -- (batch-junction);
\draw[wire] (minibatch-one.north |- batch-junction) --
    (minibatch-final.north |- batch-junction);
\fill[figureInk] (batch-junction) circle (1pt);
\foreach \step in {one,two,final} {
    \draw[edge] (minibatch-\step.north |- batch-junction) -- (minibatch-\step.north);
    \draw[trainer edge] (minibatch-\step.south) -- (update-\step);
}

\draw[edge] (training-current.east) -- (193,13) -- (193,117)
    -- node[above,detail,inner sep=1pt] {Sync for next rollout} (23,117)
    -- (execution-rollout.north);
\draw[edge,draw=figureMuted!65,densely dashed]
    (training-current.north) -- (execution-current.south);

\draw[figureNeutralEdge,line width=0.7pt] (199,-1) -- (199,173);
\node[panel heading] at (205,174) {Our method: EPRO};

\draw[card,draw=figureBlueEdge!75,fill=figureBlueTint]
    (205,74) rectangle (298,156);
\draw[card,draw=figureAmberEdge!70,fill=figureAmberTint]
    (302,74) rectangle (395,156);
\node[panel heading,text=figureBlue,text width=85pt] at (209,152)
    {Execution-policy\\prediction};
\node[panel heading,text=figureAmber,text width=85pt] at (306,152)
    {Prediction-risk\\estimation};
\node[detail,anchor=north west,text width=85pt,font=\fontsize{6.5}{7.6}\selectfont]
    at (209,132)
    {Estimate the unscored\\current execution policy\\from the rollout anchor\\
     and training-policy changes.};
\node[detail,anchor=north west,text width=85pt,font=\fontsize{6.5}{7.6}\selectfont]
    at (306,132)
    {Measure distribution-wide\\policy movement as a\\proxy for transport\\
     prediction uncertainty.};
\node[formula] at (251.5,88) {\resizebox{85pt}{!}{$\displaystyle
    \widehat{\mu}_\theta(a\mid s)\propto
    \mu_{\theta_0}(a\mid s)\frac{\pi_\theta(a\mid s)}{\pi_{\theta_0}(a\mid s)}$}};
\node[formula] at (348.5,88) {\resizebox{85pt}{!}{%
    $\begin{gathered}
    D_{\mathrm{shift}}\!\left(\pi_\theta(\cdot\mid s),
                              \pi_{\theta_0}(\cdot\mid s)\right)\\[-2pt]
    {}=\sqrt{\operatorname{Var}_{a\sim\pi_{\theta_0}(\cdot\mid s)}
        \!\left[\log\frac{\pi_\theta(a\mid s)}{\pi_{\theta_0}(a\mid s)}\right]}
    \end{gathered}$}};

\draw[card,draw=figureGradientEdge,fill=figureGradientTint]
    (205,-2) rectangle (395,65);
\draw[edge,draw=figureBlue] (251.5,74) -- (251.5,65);
\draw[edge,draw=figureAmber] (348.5,74) -- (348.5,65);
\node[panel heading,text=figureGradientInk] at (213,59) {EPRO gradient};
\node[formula,anchor=west] at (213,26) {\resizebox{174pt}{!}{%
    $\begin{aligned}
    &\nabla_\theta \mathcal L_t(\theta)
    = -A_t\,\frac{\widehat{\mu}_\theta(a_t\mid s_t)}{\mu_{\theta_0}(a_t\mid s_t)}\,
    f_{\mathrm{soft\,clip}}(D_{\mathrm{shift}})\\[1pt]
    &\quad{}\times\Bigl[\nabla_\theta\log\pi_\theta(a_t\mid s_t)
      -\mathbb E_{v\sim\widehat{\mu}_\theta(\cdot\mid s_t)}
         \!\bigl[\nabla_\theta\log\pi_\theta(v\mid s_t)\bigr]\Bigr]
    \end{aligned}$}};
\end{tikzpicture}%
\endgroup

            \vspace{0.1cm}

            \includegraphics[width=\linewidth]{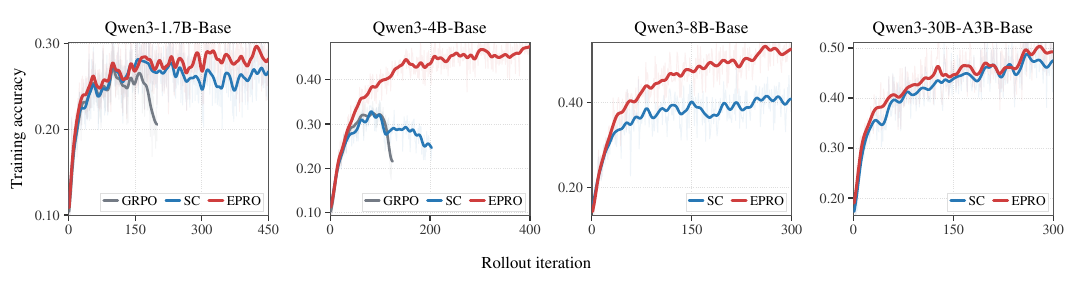}

        \end{minipage}
    \end{adjustbox}

    \caption{
    \textbf{Overview of Execution-Policy Robust Optimization (EPRO).}
    \textbf{Top left:} In decoupled LLM RL, the training policy evolves across learner updates, while the corresponding current execution policy remains unobserved without additional inference-engine evaluations.
    \textbf{Top right:} EPRO optimizes an unobserved execution policy through an imperfect predictor, accounting for prediction error and adapting updates through a robust optimization principle.
    \textbf{Bottom:} Training-accuracy trajectories across four Qwen3 model settings illustrate EPRO's improvements over the compared baselines.
}

    \label{fig:engines}
\end{figure}

Our ultimate objective, however, is to optimize the \emph{execution policy} induced by the training policy's current parameters. At the rollout checkpoint, both the training and execution policies can be observed. During subsequent learner updates, the current training policy remains accessible, but the corresponding \emph{current execution policy}---the distribution that the inference engine would induce if loaded with the training policy's latest parameters---cannot be directly observed. Obtaining it after every update would require additional synchronization, extra GPU memory and inference-engine computation. Consequently, while we know the execution policy at the rollout checkpoint and can track changes in the training policy, the current execution policy remains unobserved.

This reveals a fundamental challenge:

\begin{notebox}{accentblue}
\centering
How can we reliably optimize the current execution policy,
when its evolution under learner updates
cannot be directly observed?
\end{notebox}

To address this challenge, we introduce \textbf{Execution-Policy Robust Optimization} (EPRO), a framework that predicts the evolving execution policy and optimizes the evolving execution policy through uncertainty-aware optimization of its predicted counterpart. We develop the framework through three connected research questions, which also organize the main sections of this paper.

\textbf{RQ1: Can we predict how the execution policy evolves under learner updates?} We use a simple straight-through estimator (STE)-style approximation, estimating changes in the execution policy from the corresponding changes in the training policy, starting from the observed rollout-time execution distribution. This yields a differentiable estimate of the current execution policy, whose score leads to Transported Score Centering (TSC). However, while this provides an update direction based on execution policy, the approximation introduces non-negligible estimation errors.

\textbf{RQ2: How should we optimize with an imperfect execution-policy prediction?} We formulate a robust optimization framework to account for uncertainty in the predicted execution policy. Our analysis shows that larger discrepancies between the estimated and actual execution policies requires smaller updates to maintain a robust policy-improvement guarantee. This motivates an uncertainty-aware soft clipping, which smoothly attenuates updates according to prediction uncertainty. However, since the actual execution policy is unknown, the actual prediction error remains unobservable during learner updates.

\textbf{RQ3: Can we identify and control the prediction risk?}
We introduce Policy Shift, an observable statistic
that characterizes distribution-wide changes in the training policy.
Empirically, it identifies high transport-error positions more
effectively than a sampled-action ratio.
We use this signal to construct
Shift-Aware Soft Clipping (SSC),
which smoothly attenuates risky outward updates.
Together, TSC and SSC form EPRO:
the transported prediction provides the update direction,
while Policy Shift controls how strongly that prediction
should be trusted.

We implement EPRO and evaluate it on Qwen3-1.7B-Base, Qwen3-4B-Base, Qwen3-8B-Base, and Qwen3-30B-A3B-Base, using low-precision or quantized inference for rollout generation and BF16 for training. EPRO achieves the highest average accuracy among the compared methods across seven mathematical reasoning benchmarks on all four models.

\section{RQ1: How Should We Predict the Unobserved Execution Policy?}

\subsection{The Missing Execution Policy}

In decoupled LLM reinforcement learning, separate numerical engines for rollout generation and gradient computation may induce different policies despite sharing the same model parameters $\theta$. We denote the training policy by $\pi_\theta$ and the execution policy by $\mu_\theta$, with the objective of maximizing $J(\mu_\theta)$.

At the beginning of each rollout iteration, the training parameters are synchronized to the inference engine at checkpoint $\theta_0$, where both $\pi_0$ and $\mu_0$ can be observed. During subsequent learner updates, $\pi_\theta$ remains accessible, but the corresponding execution policy $\mu_\theta$ is unknown. Evaluating $\mu_\theta$ after every update would require repeated synchronization and additional inference-engine computation and memory. Thus, while the learner knows the execution policy at $\theta_0$ and can track the training policy's evolution, it cannot directly observe the current execution policy. We therefore seek to estimate $\mu_\theta$ from the observed rollout-time distributions and the accessible training-policy changes.

\subsection{From STE to Execution-Policy Prediction}

To estimate the evolution of the execution policy, we adopt a straight-through estimator (STE) perspective~\citep{bengio2013estimatingSTE}. Let $z_{\mu_\theta}(s)$ and $z_{\pi_\theta}(s)$ denote the centered next-token logit vectors produced by the execution and training engines, where $s$ is a response prefix. Since both numerical paths share the same model parameters, we approximate the execution-policy response to parameter updates using the corresponding response of the training policy.

Specifically, we approximate their local logit Jacobians as equal. Integrating this approximation along the parameter path from $\theta_0$ to $\theta$ gives
\begin{equation}
\nabla_\theta z_{\mu_\theta}(s) \overset{\mathrm{STE}}{\approx} \nabla_\theta z_{\pi_\theta}(s),\quad
\hat z_{\mu_\theta}(s) = z_{\mu_0}(s) + z_{\pi_\theta}(s) - z_{\pi_0}(s).
\label{eq:ste_prediction}
\end{equation}

The resulting estimate starts from the observed execution logits at the rollout checkpoint and uses the training-logit changes to approximate their subsequent evolution. Importantly, this approximation does not require the training and execution policies to coincide; it preserves their observed initial difference while borrowing the parameter response from the training engine.

Normalizing the estimated logits gives a differentiable estimate of the current execution policy. Its corresponding score follows directly by differentiation:
\begin{equation}
\begin{aligned}
\hat\mu_\theta(a\mid s) &= \frac{\mu_0(a\mid s)\,\pi_\theta(a\mid s)/\pi_0(a\mid s)}{Z_\theta(s)},\\[8pt]
\nabla_\theta\log\hat\mu_\theta(a\mid s) &= \nabla_\theta\log\pi_\theta(a\mid s) - \mathbb{E}_{v\sim\hat\mu_\theta(\cdot\mid s)}\left[\nabla_\theta\log\pi_\theta(v\mid s)\right],
\end{aligned}
\label{eq:execution_prediction}
\end{equation}
where $Z_\theta(s)=\sum_{v\in\mathcal V}\mu_0(v\mid s)\pi_\theta(v\mid s)/\pi_0(v\mid s)$.

By construction, $\hat\mu_{\theta_0}=\mu_0$. As the learner updates the model, $\hat\mu_\theta$ evolves with the training policy while retaining the execution-policy information observed at $\theta_0$. The resulting score in Eq.~\eqref{eq:execution_prediction} centers the training-policy score under the estimated execution policy. At the rollout checkpoint, it recovers conventional Score Centering (SC)\citep{marek2026score}. During subsequent updates, the centering distribution evolves with $\hat\mu_\theta$ rather than remaining fixed at $\mu_0$. We call this extension Transported Score Centering (TSC).

\subsection{Accumulation of Prediction Error}

The execution-policy estimate in Eq.~\eqref{eq:execution_prediction} relies on the STE approximation, which need not hold exactly. Differences between the local responses of the training and execution engines introduce errors into the estimated execution-policy evolution. Under the corresponding path regularity conditions, the centered-logit prediction error can be expressed as the integral of the local STE approximation error along the parameter-update path:
\begin{equation}
z_{\mu_\theta}(s)_v-\hat z_{\mu_\theta}(s)_v=\int_{\theta_0}^{\theta}\underbrace{\left(\nabla_\vartheta z_{\mu_\vartheta}(s)_v-\nabla_\vartheta z_{\pi_\vartheta}(s)_v\right)}_{\text{Local STE approximation error}}{}^{\top}\mathrm{d}\theta.
\label{eq:prediction_error}
\end{equation}

Equation~\eqref{eq:prediction_error} reveals how prediction errors arise and accumulate. At the rollout checkpoint, the prediction is exact because the execution policy is directly observed. As the learner moves away from this checkpoint, local STE approximation errors accumulate along the parameter-update path. Consequently, even small discrepancies between the local responses of the two engines can lead to substantial prediction errors over larger parameter displacements. An estimate that is accurate near its initial checkpoint may therefore become less reliable as the policy lag increases.

To examine this effect, we simulate increasing policy lag along a recorded RL update direction. Starting from the rollout checkpoint $\theta_0$, we take a saved learner update $\Delta\theta_{\mathrm{RL}}$ and scale it to obtain $\theta(\alpha)=\theta_0+\alpha\Delta\theta_{\mathrm{RL}}$. Here, $\alpha=1$ reproduces the recorded update, while larger values simulate the growing parameter displacement that would arise from successive learner updates moving in the same RL direction. At each scale, we evaluate the corresponding execution policy directly using the inference engine and compare it with the prediction obtained from Eq.~\eqref{eq:execution_prediction}. Further experimental details are provided in Appendix~C.

\begin{figure}[t]
    \centering
    \begin{subfigure}[t]{0.48\textwidth}
        \centering
        \includegraphics[width=\linewidth]{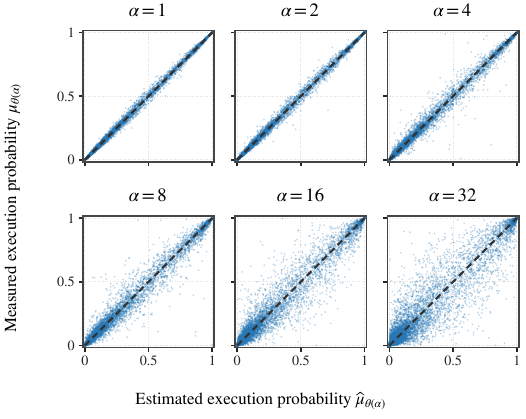}
        \caption{Estimated versus measured execution probabilities on fixed eight-token candidate sets.}
        \label{fig:execution_prediction_prob}
    \end{subfigure}
    \hfill
    \begin{subfigure}[t]{0.48\textwidth}
        \centering
        \includegraphics[width=\linewidth]{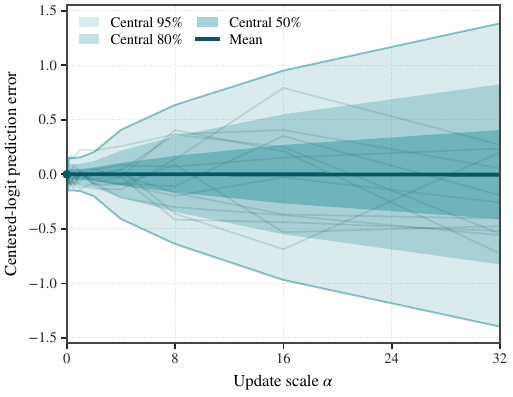}
        \caption{Centered-logit prediction residuals along the same RL update direction.}
        \label{fig:execution_prediction_residual}
    \end{subfigure}
    \caption{\textbf{Execution-policy prediction under increasing simulated policy lag.} Starting from the rollout checkpoint, we scale a recorded RL update to simulate progressively larger parameter displacements, with $\alpha=1$ corresponding to the actual recorded step. Panel (a) compares predicted and measured execution probabilities normalized over the same eight-token candidate sets. Panel (b) shows the empirical spread of centered-logit prediction residuals across 102,496 fixed token-coordinate trajectories, with shaded bands representing the central 50\%, 80\%, and 95\% ranges. \textbf{As the policy lag increases, execution-policy prediction becomes less accurate}: the estimated probabilities deviate increasingly from their measured counterparts, while the distribution of centered-logit residuals broadens substantially. These results illustrate the accumulation of prediction error as the learner moves farther from its observed execution-policy checkpoint.}
    \label{fig:execution_prediction}
\end{figure}

As shown in \cref{fig:execution_prediction_prob}, the estimated execution policy closely tracks the measured execution probabilities near the rollout checkpoint, but the agreement deteriorates as the update scale increases. \cref{fig:execution_prediction_residual} further shows that the distribution of centered-logit prediction residuals broadens with increasing parameter displacement, consistent with the accumulation of local STE approximation errors described by \cref{eq:prediction_error}. A complementary comparison with simpler predictors (\cref{app:transport-measurement}, \cref{fig:predictor-comparison}) shows that transport becomes more accurate than the frozen rollout-time execution policy at larger simulated update scales.

These results highlight a fundamental limitation of STE-based execution-policy prediction: although it provides a useful approximation of the evolving execution policy, its reliability decreases as prediction errors accumulate over larger parameter updates. This motivates the need for optimization methods that explicitly account for uncertainty in the estimated execution policy.

\section{RQ2: How Should We Optimize under Execution-Policy Prediction Error?}\label{sec:robust}

\begin{figure}[ht]
\centering
\includegraphics[width=.84\linewidth]{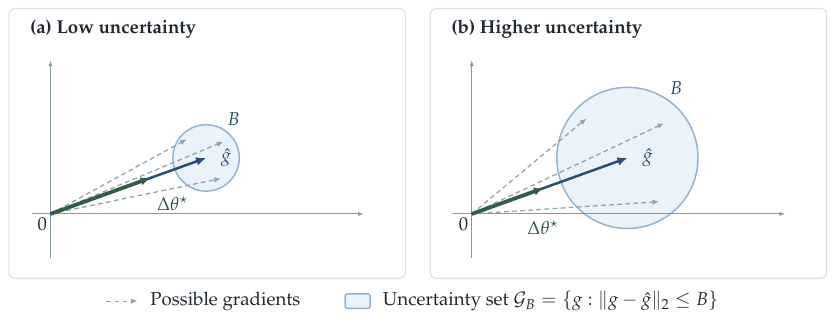}
\caption{\textbf{Gradient uncertainty and robust update attenuation.} The transported predictor provides an estimated gradient $\hat g$, while the unknown execution-policy gradient $g$ lies within a bounded uncertainty region around $\hat g$. As the uncertainty radius increases, the possible true gradients become less reliably aligned with the predicted direction, leading to a smaller robustly justified update.}
\label{fig:geometry}
\end{figure}
\subsection{A Robust Optimization View}

The transported predictor in \cref{eq:execution_prediction} provides a differentiable approximation $\widehat\mu_\theta$ of the execution policy $\mu_\theta$. While its score enables training-side optimization, the ultimate objective is to improve $J(\mu_\theta)$ rather than $J(\widehat\mu_\theta)$. This distinction matters because an update that improves the predicted policy need not improve the actual execution policy. In particular, the reliability of a surrogate ascent step depends on the alignment between the predicted and true objective gradients. Prediction errors can affect this alignment through differences in both the induced policy distributions and their responses to parameter updates. Consequently, even when $\widehat\mu_\theta$ is close to $\mu_\theta$ in distribution, its gradient may not provide a sufficiently reliable ascent direction. As illustrated in \cref{fig:geometry}, such misalignment can cause an update that improves the predicted policy to decrease the return of the actual execution policy.

Crucially, the learner cannot directly evaluate the current execution policy during intermediate updates, and therefore cannot determine how accurately the predicted gradient reflects the true improvement direction. Optimizing the transported prediction alone implicitly treats this uncertain estimate as exact. A more conservative approach is to account for the possible execution policies consistent with the prediction, and seek updates that remain beneficial despite this uncertainty. This naturally motivates a \textbf{distributionally robust optimization view}, in which the actual execution policy is considered to lie within an uncertainty neighborhood around $\widehat\mu_\theta$.

We therefore consider the possible execution policies around the transported prediction rather than optimizing that prediction alone using a distributionally robust optimization view~\citep{delage2010distributionallyDRO,ben1998robust}. For a policy distance $d$ and an uncertainty radius $\epsilon$, the distributionally robust problem is
\begin{equation}\label{eq:dro}
\max_\theta\;\min_{d(\mu,\widehat\mu_\theta)\leq\epsilon} J(\mu).
\end{equation}
The inner problem evaluates the least favorable policy within the neighborhood of $\widehat\mu_\theta$. The radius $\epsilon$ describes the uncertainty of the execution-policy prediction; it can be chosen as a quantile of the prediction-error distribution so that the neighborhood is a confidence region for the actual execution policy. 

The robust objective suggests studying how one update performs when the execution policy may occupy any point in this neighborhood. We analyze this decision in a local policy representation with an exact smooth objective $J$, isolating uncertainty in the execution policy from sampling noise in the RL estimates. The surrogate supplies the gradient at $\widehat\mu_\theta$, and a common local displacement is applied to the possible actual policies.

\subsection{A High-probability Improvement Guarantee}
\begin{theorem}[Robust policy improvement]
\label{thm:robust-improvement}
Suppose the execution-policy objective $J(\mu_\theta)$ is
$L$-smooth with respect to model parameters, with $L>0$,
and the local STE approximation error is bounded by $\rho$.

Suppose the predicted execution policy satisfies
\begin{equation}
\Pr\left(
D_{\mathrm{KL}}(\mu_\theta\|\hat\mu_\theta)
\leq \varepsilon_{\mathrm{pred}}
\right)
\geq 1-\delta.
\label{eq:prediction-confidence}
\end{equation}

Under the bounded-score and finite-horizon conditions
specified in \cref{app:robust-proof},
let
$\hat g=\nabla_\theta J(\hat\mu_\theta)$.
Then, with probability at least $1-\delta$, the update
\begin{equation}
\theta^+
=
\theta+
\frac{1}{L}
\left(
1-
\frac{
C_1\sqrt{\varepsilon_{\mathrm{pred}}}+C_2\rho
}{
\|\hat g\|_2
}
\right)_+
\hat g
\label{eq:robust-update}
\end{equation}
satisfies
\begin{equation}
J(\mu_{\theta^+})-J(\mu_\theta)
\geq
\frac{
\left(
\|\hat g\|_2
-
C_1\sqrt{\varepsilon_{\mathrm{pred}}}
-
C_2\rho
\right)_+^2
}{2L}
\geq 0.
\label{eq:robust-guarantee}
\end{equation}

Here $C_1,C_2$ are constants specified in
\cref{app:robust-proof}.
When the surrogate gradient vanishes, the update is zero.
\end{theorem}
The proof is provided in \cref{app:robust-proof}. \cref{thm:robust-improvement} suggests two practical requirements for optimizing with an imperfect execution-policy predictor.

\textbf{Finding a proxy for prediction error.} The uncertainty radius $\epsilon_{\mathrm{pred}}$ bounds the discrepancy between the predicted and actual execution policies, which in turn bounds the error in the predicted gradient. However, the true prediction error cannot be directly observed during learner updates, since doing so would require evaluating the current execution policy. We therefore need an observable training-side proxy that can identify where execution-policy predictions are less reliable.

\textbf{Controlling updates through soft clipping.} \cref{thm:robust-improvement} shows that, with the predicted gradient fixed, the robustly justified update magnitude decreases continuously as prediction uncertainty increases. This motivates an uncertainty-aware soft-clipping mechanism that smoothly attenuates risky updates while allowing stronger updates where predictions are more reliable. The following section develops an observable risk proxy and uses it to construct such a mechanism.

\section{RQ3: Can We Identify and Control Prediction Risk?}\label{sec:risk}
The robust-improvement result motivates reducing update strength when the execution-policy prediction is less reliable. Since the true prediction error is unavailable during learner updates, we first seek a measurable training-side proxy and then use it to control the update itself.

\subsection{Distributional Policy Change as an Observable Risk Signal}\label{sec:shift-evidence}
A sampled-action ratio describes the probability change of one token, whereas the transport predictor responds to changes across the next-token distribution. We therefore consider the variability of log-probability changes under the rollout-time training policy:
\begin{equation}\label{eq:shift}
\operatorname{Shift}_\theta(s)
=\sqrt{\operatorname{Var}_{a\sim\pi_0(\cdot\mid s)}\!\left[\log\frac{\pi_\theta(a\mid s)}{\pi_0(a\mid s)}\right]}.
\end{equation}
We call this statistic \emph{Policy Shift}. Unlike a sampled-action ratio, it measures movement across the entire next-token distribution. This is equivalently the spread of training-side logit changes:
\begin{equation}\label{eq:shift-logit}
\operatorname{Shift}_\theta(s)^2
=\operatorname{Var}_{a\sim\pi_0(\cdot\mid s)}
\!\left[z_{\pi_\theta}(s)_a-z_{\pi_0}(s)_a\right].
\end{equation}
The equality holds because the difference between a log-probability change and a logit change is constant across tokens, and variance is unaffected by a constant offset.

We test whether this observable statistic identifies inaccurate transport predictions using execution-engine replay. At each update scale, we label the highest-error 10\% of held-out outward positions according to measured execution-versus-transport KL on common candidate support. We compare Policy Shift with a sampled-action ratio score on this ranking task; the precise evaluation protocol is in \cref{app:risk-measurement}.



\begin{figure}[!htbp]
    \centering
    \begin{subfigure}[t]{0.38\linewidth}
        \centering
        \includegraphics[width=\linewidth]{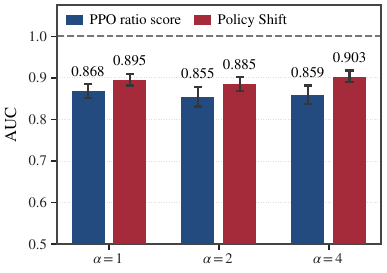}
        \caption{Detecting high transport-error positions. The positive class is the top 10\% of measured common-support transport KL within each update scale. Policy Shift is compared with the PPO sampled-action ratio; uncertainty intervals are detailed in \cref{app:risk-measurement}.}
        \label{fig:auc}
    \end{subfigure}
    \hfill
    \begin{subfigure}[t]{0.58\linewidth}
        \centering
        \includegraphics[width=\linewidth]{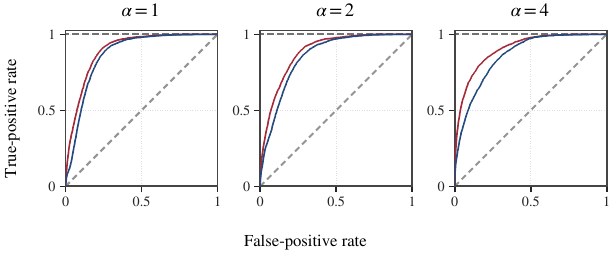}
        \caption{ROC curves for the same high-error classification task. The distributional signal ranks more high-transport-error positions above low-error positions across the displayed operating ranges. The dashed diagonal indicates random ranking.}
        \label{fig:roc}
    \end{subfigure}
\end{figure}

\noindent\textbf{Policy Shift identifies unreliable predictions.} Across all three tested update scales, it ranks the high transport-error positions more accurately than the sampled-action ratio (\cref{fig:auc,fig:roc}). At a given false-positive rate, the distributional signal captures more of the positions where transport is least accurate. The theorem motivates smaller updates under greater uncertainty; this measurement gives an observable way to recognize that risk.

\FloatBarrier

\subsection{From Prediction Risk to EPRO}
\label{sec:method}

\begin{wrapfigure}{r}{0.25\textwidth}
\vspace{-10pt}
\centering
\includegraphics[width=\linewidth]{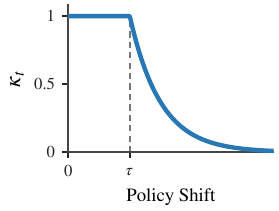}
\caption{Smooth attenuation function.}
\label{fig:gate}
\vspace{-8pt}
\end{wrapfigure}

EPRO combines two components already motivated above: \emph{Transported Score Centering} (TSC) follows the predicted execution policy, while \emph{Shift-Aware Soft Clipping} (SSC) reduces outward updates where that prediction is less reliable.

\paragraph{Shift-Aware Soft Clipping.}
We use Policy Shift to attenuate updates that move the sampled action further in the direction favored by its advantage:
\begin{equation}\label{eq:gate}
\kappa_t =
\begin{cases}
\displaystyle \exp\!\left(-\frac{[\operatorname{Shift}_\theta(s_t)-\tau]_+}{\sigma}\right),
& \displaystyle A_t\log\frac{\pi_\theta(a_t\mid s_t)}{\pi_0(a_t\mid s_t)}>0,\\[5pt]
1,&\text{otherwise}.
\end{cases}
\end{equation}
Here $\tau$ sets the onset of attenuation and $\sigma>0$ its strength.

The sampled-action likelihood ratio determines \emph{whether} the update is outward; Policy Shift determines \emph{how strongly} it is attenuated (\cref{fig:gate}). Non-outward contributions are retained.

\paragraph{The EPRO update.}
Combining the transported score with SSC gives a compact surrogate loss:
\begin{equation}\label{eq:loss}
\mathcal L_{\mathrm{EPRO}}(\theta)
=-\E_t\!\left[
\sg\!\left(A_t\min\{\frac{\widehat\mu_\theta(a_t\mid s_t)}{\mu_0(a_t\mid s_t)},C\}\cdot\kappa_t\right)
\log\widehat\mu_\theta(a_t\mid s_t)
\right].
\end{equation}
Here $C$ caps the importance ratio, and $\sg$ denotes stop-gradient. Its update direction is
\begin{align}
&-\nabla_\theta\mathcal L_{\mathrm{EPRO}}=\E_t\Biggl[
A_t\,
\underbrace{\min\!\left\{\frac{\widehat\mu_\theta(a_t\mid s_t)}{\mu_0(a_t\mid s_t)},\,C\right\}}_{\substack{\text{TIS between sampling}\\\text{and predicted distribution}}}
\;\underbrace{\kappa_t}_{\substack{\text{Prediction-risk-aware}\\\text{attenuation}}}\notag\\[9pt]
&\qquad\cdot\underbrace{\left(
\nabla_\theta\log\pi_\theta(a_t\mid s_t)
-\E_{v\sim\widehat\mu_\theta(\cdot\mid s_t)}\!\left[\nabla_\theta\log\pi_\theta(v\mid s_t)\right]
\right)}_{\substack{\text{Transported score centering}\\\text{for the predicted execution policy}}}
\Biggr].\label{eq:epro_gradient}
\end{align}
The importance weight and SSC attenuation are held fixed during differentiation.

EPRO does not require additional execution-engine evaluations during learner updates. At rollout time, we record a top-$K$ execution distribution and combine the remaining vocabulary into a tail bucket. Subsequent trainer forward passes approximate the transport normalizer, score-centering correction, and Policy Shift; details are given in \cref{app:implementation}.

\section{Experiments}\label{sec:experiments}
\subsection{Experimental Setup}
\noindent\textbf{Training.} We train Qwen3-1.7B-Base, Qwen3-4B-Base, Qwen3-8B-Base, and Qwen3-30B-A3B-Base with low-precision or quantized execution engines for rollouts and BF16 training~\citep{yang2025qwen3}. Each rollout iteration is followed by several learner minibatch updates. Model-specific training settings are collected in \cref{app:training-config}.

\noindent\textbf{Evaluation.} We evaluate our models on seven math benchmarks, including MATH500~\citep{lightman2024letMATH500}, Minerva~\citep{lewkowycz2022solvingMINERVA}, Olympiad~\citep{he2024olympiadbench}, AIME 2024~\citep{aime24}, AIME 2025~\citep{aime25}, AIME 2026~\citep{aime26}, and AMC 2023. We use mean@8 for the first three benchmarks and mean@32 for the other four.

\noindent\textbf{Baselines.} We compare against GRPO, SC, SC + PPO gate, and TSC + PPO gate. Here SC uses the rollout execution distribution for score centering, while the PPO gate controls outward updates using a sampled-action ratio. TSC + PPO gate replaces SC with transported centering under the same gate, isolating the contribution of transport. EPRO replaces that ratio gate with shift-aware attenuation. The early-terminated collapsed runs are included in the training curves, and the update formulations are summarized in \cref{app:baselines}.

\subsection{Main Results}
We examine both the progress of RL training and the reasoning performance of the resulting checkpoints. The training curves capture how each method behaves over successive updates, including runs that terminate early; held-out evaluation then shows how these differences carry over to mathematical reasoning tasks.

\begin{figure}[t]
\centering
\includegraphics[width=.98\linewidth]{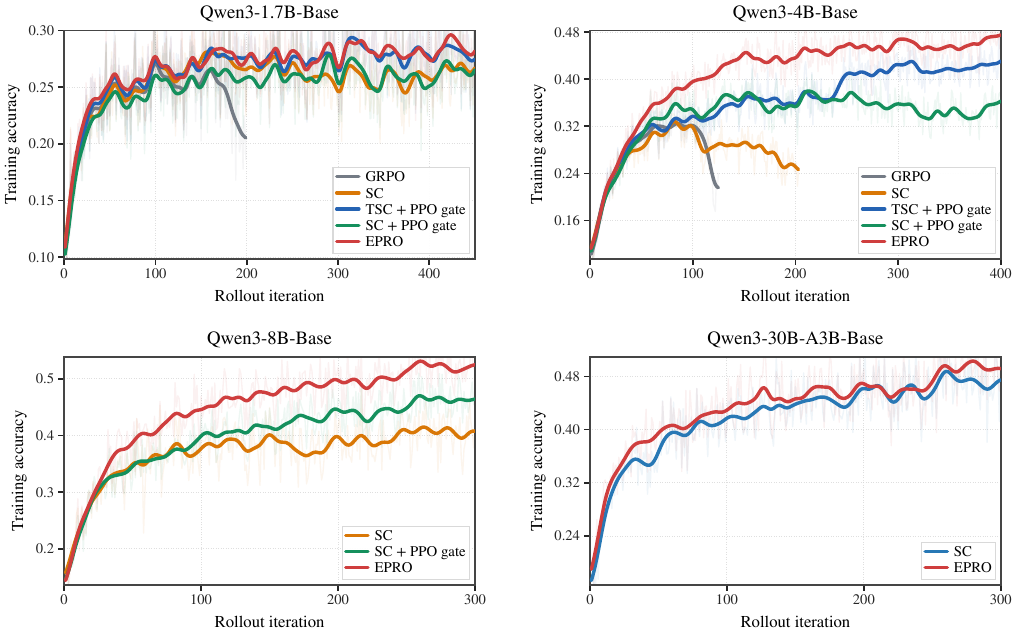}
\caption{Training accuracy with low-precision rollout execution and BF16 training, smoothed over ten actor optimizer steps. The plotted score is the mean binary correctness reward; short trajectories indicate terminated runs.}
\label{fig:training}
\end{figure}

\noindent\textbf{Training accuracy.} As shown in \cref{fig:training}, all methods improve rapidly at the beginning, but their trajectories diverge as training continues. EPRO stays among the strongest runs on 1.7B and 30B-A3B, while building a clearer lead on 4B and 8B, particularly in later training. The figure also shows runs that fail to sustain this progress: GRPO declines and terminates early on 1.7B and 4B, and SC on 4B also deteriorates before termination. The resulting picture is not only one of higher accuracy, but of stronger performance over sustained RL updates.

\begin{table}[t]
\centering
\caption{Held-out mathematical reasoning accuracy (\%; higher is better). Values are means with standard deviations over three evaluation seeds. Bold indicates the highest mean for each benchmark within a model group; Avg. is the unweighted mean across all seven benchmarks.}
\label{tab:math-eval}
\begingroup
\newcommand{\evalscore}[2]{$#1_{\scriptscriptstyle\pm#2}$}
\setlength{\tabcolsep}{3.3pt}
\renewcommand{\arraystretch}{1.12}
\small
\resizebox{\textwidth}{!}{%
\begin{tabular}{@{}lcccccccc@{}}
\toprule
\textbf{Method} & \textbf{MATH500} & \textbf{Minerva} & \textbf{Olympiad} & \textbf{AIME24} & \textbf{AIME25} & \textbf{AIME26} & \textbf{AMC23} & \textbf{Avg.} \\
\midrule
\multicolumn{9}{@{}l}{\textbf{Qwen3-1.7B-Base}} \\
\addlinespace[2pt]
Base model& \evalscore{24.3}{0.6} & \evalscore{4.4}{0.3} & \evalscore{8.3}{0.3} & \evalscore{0.9}{0.2} & \evalscore{0.8}{0.2} & \evalscore{0.8}{0.2} & \evalscore{13.8}{0.5} & 7.6 \\
SC & \evalscore{\mathbf{57.1}}{0.4} & \evalscore{16.3}{0.4} & \evalscore{\mathbf{24.7}}{0.4} & \evalscore{6.0}{0.4} & \evalscore{3.2}{0.5} & \evalscore{2.2}{0.4} & \evalscore{37.2}{1.2} & 21.0 \\
SC + PPO gate & \evalscore{52.5}{0.4} & \evalscore{17.6}{0.4} & \evalscore{22.3}{0.4} & \evalscore{4.6}{0.3} & \evalscore{3.1}{0.5} & \evalscore{2.1}{0.1} & \evalscore{33.6}{0.9} & 19.4 \\
TSC + PPO gate & \evalscore{53.1}{0.4} & \evalscore{16.6}{0.5} & \evalscore{22.5}{0.3} & \evalscore{4.0}{0.4} & \evalscore{\mathbf{3.8}}{0.1} & \evalscore{\mathbf{3.1}}{0.9} & \evalscore{\mathbf{39.5}}{0.4} & 20.4 \\
\rowcolor{eproblue}
\textbf{EPRO} & \evalscore{55.3}{0.3} & \evalscore{\mathbf{17.9}}{0.6} & \evalscore{23.3}{0.3} & \evalscore{\mathbf{11.9}}{0.3} & \evalscore{3.4}{0.7} & \evalscore{2.6}{0.9} & \evalscore{35.7}{0.8} & \textbf{21.5} \\
\midrule
\multicolumn{9}{@{}l}{\textbf{Qwen3-4B-Base}} \\
\addlinespace[2pt]
Base model & \evalscore{34.2}{0.3} & \evalscore{9.0}{1.0} & \evalscore{16.6}{0.3} & \evalscore{5.2}{0.9} & \evalscore{3.1}{0.4} & \evalscore{3.0}{1.3} & \evalscore{24.3}{1.2} & 13.6 \\
SC + PPO gate & \evalscore{65.8}{0.6} & \evalscore{27.3}{0.1} & \evalscore{36.8}{0.4} & \evalscore{13.6}{1.6} & \evalscore{11.3}{0.6} & \evalscore{11.8}{0.3} & \evalscore{53.4}{1.4} & 31.4 \\
TSC + PPO gate & \evalscore{67.2}{0.4} & \evalscore{26.7}{0.2} & \evalscore{38.6}{0.2} & \evalscore{\mathbf{17.6}}{0.5} & \evalscore{16.9}{0.6} & \evalscore{11.5}{0.6} & \evalscore{57.4}{0.5} & 33.7 \\
\rowcolor{eproblue}
\textbf{EPRO} & \evalscore{\mathbf{70.7}}{0.2} & \evalscore{\mathbf{28.2}}{0.1} & \evalscore{\mathbf{41.7}}{0.1} & \evalscore{17.5}{0.3} & \evalscore{\mathbf{20.5}}{1.4} & \evalscore{\mathbf{15.1}}{0.9} & \evalscore{\mathbf{63.3}}{0.7} & \textbf{36.7} \\
\midrule
\multicolumn{9}{@{}l}{\textbf{Qwen3-8B-Base}} \\
\addlinespace[2pt]
Base model & \evalscore{41.6}{0.9} & \evalscore{8.8}{0.5} & \evalscore{18.8}{0.5} & \evalscore{6.7}{0.4} & \evalscore{4.8}{0.5} & \evalscore{4.3}{0.1} & \evalscore{32.1}{0.5} & 16.7 \\
SC & \evalscore{67.7}{0.3} & \evalscore{26.5}{0.5} & \evalscore{37.4}{0.3} & \evalscore{20.2}{0.5} & \evalscore{13.9}{1.3} & \evalscore{14.5}{0.4} & \evalscore{65.7}{0.9} & 35.1 \\
SC + PPO gate & \evalscore{70.0}{0.4} & \evalscore{28.9}{0.2} & \evalscore{41.2}{0.4} & \evalscore{17.5}{1.1} & \evalscore{16.7}{0.5} & \evalscore{16.6}{0.7} & \evalscore{64.2}{1.1} & 36.5 \\
\rowcolor{eproblue}
\textbf{EPRO} & \evalscore{\mathbf{73.7}}{0.4} & \evalscore{\mathbf{29.9}}{0.4} & \evalscore{\mathbf{44.5}}{0.2} & \evalscore{\mathbf{21.6}}{1.1} & \evalscore{\mathbf{18.4}}{0.7} & \evalscore{\mathbf{18.6}}{1.3} & \evalscore{\mathbf{67.1}}{0.4} & \textbf{39.1} \\
\midrule
\multicolumn{9}{@{}l}{\textbf{Qwen3-30B-A3B-Base}} \\
\addlinespace[2pt]
Base model & \evalscore{44.2}{0.7} & \evalscore{9.1}{0.4} & \evalscore{20.0}{0.6} & \evalscore{9.5}{0.3} & \evalscore{3.5}{0.4} & \evalscore{5.6}{0.8} & \evalscore{34.6}{0.8} & 18.1 \\
SC & \evalscore{70.3}{0.5} & \evalscore{26.4}{0.2} & \evalscore{39.9}{0.5} & \evalscore{21.8}{0.8} & \evalscore{\mathbf{19.5}}{0.4} & \evalscore{14.8}{1.4} & \evalscore{67.7}{0.6} & 37.2 \\
\rowcolor{eproblue}
\textbf{EPRO} & \evalscore{\mathbf{72.3}}{0.7} & \evalscore{\mathbf{26.9}}{0.5} & \evalscore{\mathbf{41.5}}{0.2} & \evalscore{\mathbf{24.5}}{1.5} & \evalscore{19.4}{0.7} & \evalscore{\mathbf{17.0}}{0.7} & \evalscore{\mathbf{71.9}}{0.2} & \textbf{39.1} \\
\bottomrule
\end{tabular}%
}
\endgroup
\end{table}

\noindent\textbf{Evaluations.} \Cref{tab:math-eval} reports the accuracy of the evaluated checkpoints across seven mathematical reasoning benchmarks. EPRO achieves the highest average score in every model setting, from 1.7B through 30B-A3B. The gains are most pronounced on 4B and 8B, where the stronger training trajectories are matched by clear improvements on held-out tasks. Across the benchmarks, the results show that EPRO's training advantage is reflected in the reasoning performance of the final models.

\subsection{Ablations and Discussion}
\noindent\textbf{Transported Score Centering improves upon conventional centering.} On both 1.7B and 4B, replacing SC with TSC while keeping the PPO gate fixed improves held-out accuracy (\cref{tab:math-eval}), while TSC results in a higher training accuracy. On 4B, the SC run without clipping also collapses, underscoring the importance of regulating updates alongside score centering.

\smallskip
\noindent\textbf{Shift-Aware Soft Clipping provides further gains.} Holding TSC fixed, replacing the PPO ratio gate with SSC improves the held-out average on both 1.7B and 4B. The advantage is most visible on 4B: EPRO separates from TSC + PPO gate throughout later-stage training and finishes with higher reasoning accuracy. This directly connects the prediction-risk signal in \cref{sec:shift-evidence} to improved RL optimization.

\smallskip
\noindent\textbf{EPRO sustains stronger RL optimization across model scales.} On 8B, EPRO outperforms both SC and SC + PPO gate in training accuracy and held-out evaluation. On 30B-A3B, it also improves over SC. Together with the controlled comparisons on smaller models, these results demonstrate the benefit of combining execution-policy transport with prediction-aware update control.

\FloatBarrier
\section{Related Work}\label{sec:related}

Modern LLM reinforcement learning systems typically use separate inference and training engines for rollout generation and policy updates, which can induce discrepancies between the execution and training policies even when model parameters are synchronized. Related studies have investigated approximate gradient estimation, numerical precision and quantization, efficient inference implementations, and token-level distribution shifts during RL fine-tuning \citep{jacob2018quantizationandtraining,micikevicius2017mixedtraining,micikevicius2022fp8DL,xiao2023smoothquant,frantar2022gptq,li2024evaluatingquant,dao2022flashattention,meng2026sparse}. Recent studies have investigated the sources and consequences of this training–inference mismatch \citep{yao2025rolloutTIS, zhong2026diagnosingTIM}, while others address it through numerical alignment, quantization-aware training, or importance-sampling corrections \citep{qi2025defeatingFP16,gu2026qarl,yu2026rethinkingCIS}. Related efforts also investigate stable learning under policy staleness and other forms of off-policy data collection \citep{ritter2026llmsOAPL,zheng2025stabilizingRL,shen2026vespo}. These works establish the importance of handling discrepancies between behavior and training policies, primarily through mismatch reduction or off-policy correction.

Most closely related to our work, Score Centering identifies a gradient bias induced by training–inference mismatch and introduces a centering correction to stabilize policy updates \citep{marek2026score}. Subsequent analysis further distinguishes numerical discrepancies from policy lag and studies their corresponding correction mechanisms \citep{li2026originTIM,bai2026regularizedRSC}. From a complementary perspective, MIPI argues that improving the training policy does not necessarily improve the execution policy, motivating optimization directly toward execution-policy improvement \citep{liang2026mirageMIPI}. Our work shares this objective-level motivation but takes a different approach: we construct a differentiable estimate of the execution policy from observed rollouts and training-policy changes, derive its corresponding score, and explicitly account for estimation errors during optimization.

Stable policy optimization has traditionally relied on trust-region constraints and importance-ratio clipping \citep{schulman2015trustTRPO,shao2024deepseekmathGRPO,schulman2017proximal,wang2020trulyPPO}, with recent LLM-specific methods further improving update stability through revised clipping rules, sequence-level objectives, and off-policy weighting \citep{yu2026dapo,zheng2025gspo,shen2026vespo,chen2025minimaxCISPO,gao2025softSAPO,qi2026rethinkingDPPO}. Among them, DPPO replaces sampled-action ratio constraints with distribution-level divergence measures, while SAPO introduces smooth attenuation of policy updates as an alternative to hard clipping \citep{qi2026rethinkingDPPO,gao2025softSAPO}. Although our method also employs smooth attenuation and distribution-level information, its motivation is different: our robust optimization analysis relates reliable execution-policy improvement to execution-policy estimation error. This motivates Shift-Aware Soft Clipping, which uses an observable Policy Shift statistic as an empirical proxy for estimation uncertainty to regulate potentially unreliable updates.

\section{Conclusion}\label{sec:conclusion}
The goal of decoupled LLM reinforcement learning is to improve the policy that actually generates responses, even when learner gradients come from a different numerical engine. EPRO pursues this goal by transporting training-policy changes to predict the execution policy, then moderating updates when that prediction is less reliable. Transported Score Centering and Shift-Aware Soft Clipping turn these two ideas into a practical learner. Across four Qwen3 model settings, EPRO improves held-out mathematical reasoning accuracy while requiring no repeated execution-engine evaluation during learner updates.

\bibliographystyle{colm2026_conference}
\bibliography{references}

\appendix

\section{Proof of \texorpdfstring{\cref{thm:robust-improvement}}{the robust policy improvement theorem}}
\label{app:robust-proof}

We first state the assumptions used in the analysis.

\paragraph{Assumption A.1 (Local STE approximation).}
The discrepancy between the execution-policy and
training-policy logit Jacobians is bounded by $\rho$:
\begin{equation}
\left\|
\nabla_\theta z_{\mu_\theta}(s)_v
-
\nabla_\theta z_{\pi_\theta}(s)_v
\right\|_2
\leq \rho.
\label{eq:ste-assumption}
\end{equation}

\paragraph{Assumption A.2 (Prediction uncertainty).}
The transported prediction satisfies the high-probability
KL bound in Eq.~\eqref{eq:prediction-confidence}.
For the full-return analysis, the confidence event is
assumed to hold uniformly over the relevant prefixes.

\paragraph{Assumption A.3 (Regularity).}
The training-policy score is bounded by
\begin{equation}
\|\nabla_\theta\log\pi_\theta(v|s)\|_2
\leq G.
\label{eq:score-bound}
\end{equation}
Rewards satisfy $R(\tau)\in[0,1]$, and the rollout
length is at most $H$. The execution-policy objective
is differentiable and locally $L$-smooth with respect
to model parameters.

\begin{proof}
By the softmax score identity and Assumption A.1,
the actual execution-policy score satisfies
\begin{equation}
\begin{aligned}
\nabla_\theta\log\mu_\theta(a|s)
={}&
\nabla_\theta\log\pi_\theta(a|s)\\
&-
\mathbb E_{v\sim\mu_\theta}
[\nabla_\theta\log\pi_\theta(v|s)]
+r_\theta(a,s),
\end{aligned}
\label{eq:true-execution-score}
\end{equation}
where the local STE residual satisfies
\begin{equation}
\|r_\theta(a,s)\|_2\leq 2\rho.
\label{eq:ste-residual}
\end{equation}

Meanwhile, the transported score from Eq.~(3) is
\begin{equation}
\begin{aligned}
\nabla_\theta\log\hat\mu_\theta(a|s)
={}&
\nabla_\theta\log\pi_\theta(a|s)\\
&-
\mathbb E_{v\sim\hat\mu_\theta}
[\nabla_\theta\log\pi_\theta(v|s)].
\end{aligned}
\label{eq:transported-score-proof}
\end{equation}

Subtracting Eqs.~\eqref{eq:true-execution-score}
and~\eqref{eq:transported-score-proof}, the sampled-action
score terms cancel, leaving only the discrepancy between
the centering distributions and the STE residual.
Using Assumption A.3,
\begin{equation}
\begin{aligned}
&
\left\|
\nabla_\theta\log\mu_\theta(a|s)
-
\nabla_\theta\log\hat\mu_\theta(a|s)
\right\|_2\\
&\leq
G\|\mu_\theta-\hat\mu_\theta\|_1
+
2\rho.
\end{aligned}
\label{eq:score-error-l1}
\end{equation}

By Pinsker's inequality,
\begin{equation}
\|\mu_\theta-\hat\mu_\theta\|_1
\leq
\sqrt{
2D_{\mathrm{KL}}(\mu_\theta\|\hat\mu_\theta)
}.
\label{eq:pinsker}
\end{equation}

Therefore, on the confidence event in Assumption A.2,
\begin{equation}
\begin{aligned}
&
\left\|
\nabla_\theta\log\mu_\theta(a|s)
-
\nabla_\theta\log\hat\mu_\theta(a|s)
\right\|_2\\
&\leq
G\sqrt{2\varepsilon_{\mathrm{pred}}}
+
2\rho.
\end{aligned}
\label{eq:transported-score-error}
\end{equation}

We next bound the discrepancy between the gradients
of the expected-return objectives.

By the policy-gradient identity,
\begin{equation}
\nabla_\theta J(\mu_\theta)
=
\mathbb E_{\tau\sim\mu_\theta}
\left[
R(\tau)
\sum_t
\nabla_\theta\log\mu_\theta(a_t|s_t)
\right],
\label{eq:true-policy-gradient}
\end{equation}
and similarly,
\begin{equation}
\nabla_\theta J(\hat\mu_\theta)
=
\mathbb E_{\tau\sim\hat\mu_\theta}
\left[
R(\tau)
\sum_t
\nabla_\theta\log\hat\mu_\theta(a_t|s_t)
\right].
\label{eq:predicted-policy-gradient}
\end{equation}

Adding and subtracting the expectation of the
transported score under the actual execution-policy
trajectory distribution gives
\begin{equation}
\begin{aligned}
&
\nabla_\theta J(\mu_\theta)
-
\nabla_\theta J(\hat\mu_\theta)\\
={}&
\mathbb E_{\tau\sim\mu_\theta}
\left[
R(\tau)\sum_t
\left(
\nabla_\theta\log\mu_\theta(a_t|s_t)
-
\nabla_\theta\log\hat\mu_\theta(a_t|s_t)
\right)
\right]\\
&+
\left(
\mathbb E_{\tau\sim\mu_\theta}
-
\mathbb E_{\tau\sim\hat\mu_\theta}
\right)
\left[
R(\tau)\sum_t
\nabla_\theta\log\hat\mu_\theta(a_t|s_t)
\right].
\end{aligned}
\label{eq:gradient-error-decomposition}
\end{equation}

The first term captures score approximation error.
Since $R(\tau)\leq 1$ and the rollout length is
at most $H$, Eq.~\eqref{eq:transported-score-error}
bounds its norm by
\begin{equation}
H\left(
G\sqrt{2\varepsilon_{\mathrm{pred}}}
+
2\rho
\right).
\label{eq:score-contribution}
\end{equation}

The second term captures the discrepancy between
the trajectory distributions induced by the actual
and transported policies.

By Pinsker's inequality, the per-prefix total
variation distance is at most
\begin{equation}
D_{\mathrm{TV}}
(\mu_\theta,\hat\mu_\theta)
\leq
\frac{1}{2}
\sqrt{2\varepsilon_{\mathrm{pred}}}.
\end{equation}

Coupling the two autoregressive policies at each
generation step, their trajectories can differ only
if at least one coupled token differs. By the union
bound over at most $H$ generation steps,
\begin{equation}
D_{\mathrm{TV}}
(P_{\mu_\theta},P_{\hat\mu_\theta})
\leq
\frac{H}{2}
\sqrt{2\varepsilon_{\mathrm{pred}}}.
\label{eq:trajectory-tv}
\end{equation}

Furthermore, by the transported score identity
and Assumption A.3,
\begin{equation}
\|\nabla_\theta\log\hat\mu_\theta(a|s)\|_2
\leq 2G.
\label{eq:transported-score-bound}
\end{equation}

Consequently, the norm of the reward-weighted
trajectory score is bounded by $2GH$.
Using the total variation bound on the difference
of expectations, the second term in
Eq.~\eqref{eq:gradient-error-decomposition}
is bounded by
\begin{equation}
2GH^2\sqrt{2\varepsilon_{\mathrm{pred}}}.
\label{eq:distribution-contribution}
\end{equation}

Combining Eqs.~\eqref{eq:score-contribution}
and~\eqref{eq:distribution-contribution} yields
\begin{equation}
\begin{aligned}
&
\left\|
\nabla_\theta J(\mu_\theta)
-
\nabla_\theta J(\hat\mu_\theta)
\right\|_2\\
&\leq
GH(2H+1)
\sqrt{2\varepsilon_{\mathrm{pred}}}
+
2H\rho.
\end{aligned}
\label{eq:policy-gradient-error}
\end{equation}

Thus, the constants in \cref{thm:robust-improvement}
can be chosen as
\begin{equation}
C_1=\sqrt{2}\,GH(2H+1),
\qquad
C_2=2H.
\label{eq:robust-constants}
\end{equation}

Let
\[
g=\nabla_\theta J(\mu_\theta),
\qquad
\hat g=\nabla_\theta J(\hat\mu_\theta).
\]

By $L$-smoothness, for any feasible local parameter
displacement $u$,
\begin{equation}
J(\mu_{\theta+u})-J(\mu_\theta)
\geq
\langle g,u\rangle
-
\frac{L}{2}\|u\|_2^2.
\label{eq:smoothness-bound}
\end{equation}

Applying Eq.~\eqref{eq:policy-gradient-error}
and the Cauchy--Schwarz inequality,
\begin{equation}
\begin{aligned}
J(\mu_{\theta+u})-J(\mu_\theta)
\geq{}&
\langle\hat g,u\rangle\\
&-
\left(
C_1\sqrt{\varepsilon_{\mathrm{pred}}}
+
C_2\rho
\right)\|u\|_2\\
&-
\frac{L}{2}\|u\|_2^2.
\end{aligned}
\label{eq:robust-lower-bound}
\end{equation}

For a fixed update norm, the right-hand side
is maximized when $u$ is aligned with $\hat g$.
Maximizing the resulting scalar quadratic gives
\begin{equation}
u^*
=
\frac{1}{L}
\left(
1-
\frac{
C_1\sqrt{\varepsilon_{\mathrm{pred}}}
+
C_2\rho
}{
\|\hat g\|_2
}
\right)_+
\hat g.
\label{eq:optimal-robust-update}
\end{equation}

Substituting this update into
Eq.~\eqref{eq:robust-lower-bound},
\begin{equation}
J(\mu_{\theta+u^*})-J(\mu_\theta)
\geq
\frac{
\left(
\|\hat g\|_2
-
C_1\sqrt{\varepsilon_{\mathrm{pred}}}
-
C_2\rho
\right)_+^2
}{2L}
\geq 0.
\label{eq:final-improvement}
\end{equation}

The result holds on the prediction-confidence
event specified in Assumption A.2, which has
probability at least $1-\delta$.
This completes the proof.
\end{proof}

\section{Head--Tail Implementation}\label{app:implementation}
Let $q=\mu_0$, $p=\pi_0$, and $u=\pi_\theta$, and let $H$ be the old execution top-$K$ token set recorded at rollout. Approximate the execution tail by a rescaled old training tail:
\begin{equation}
\widetilde q(v)=\rho_0p(v)\quad(v\notin H),\qquad \rho_0=\frac{1-q(H)}{1-p(H)}.
\end{equation}
The resulting transport normalizer and sampled-token importance weight are
\begin{align}
\widetilde Z_\theta&=\sum_{v\in H}q(v)\frac{u(v)}{p(v)}+\rho_0\bigl(1-u(H)\bigr),\\
\widetilde w_t&=\frac{u(a_t)/p(a_t)}{\widetilde Z_\theta}.
\end{align}
For $v\in H$, define $\widetilde\mu_\theta(v)=q(v)u(v)/(p(v)\widetilde Z_\theta)$. The score correction used in \cref{eq:loss} is
\begin{equation}
C_{\mathrm{tr},t}=\sum_{v\in H}\sg\!\left[\widetilde\mu_\theta(v)-\frac{\rho_0u(v)}{\widetilde Z_\theta}\right]\log u(v).
\end{equation}
Its gradient is the head--tail transported centering term. For Policy Shift, define buckets $\mathcal B=H\cup\{\mathrm{tail}\}$, with old/current training masses $p_b,u_b$. Setting $d_b=\log(u_b/p_b)$ gives
\begin{equation}
\widehat S_{\theta,K}(s)=\left[\sum_{b\in\mathcal B}p_b\left(d_b-\sum_{c\in\mathcal B}p_cd_c\right)^2\right]^{1/2},\qquad K=32.
\end{equation}
The tail bucket makes this a coarse-grained version of \cref{eq:shift}. The rollout engine records the old execution head once; the subsequent trainer minibatches compute the weights and gate from their existing forward pass.

\paragraph{Illustrative PyTorch implementation.}
The following implementation summarizes the core EPRO update.
Here $q=\mu_0$, $p_0=\pi_0$, and $u=\pi_\theta$.
The subscript $H$ denotes the rollout-time top-$K$ tokens,
with the remaining probability mass represented by a tail
bucket. All log-probabilities are normalized over the
full vocabulary.

\begin{lstlisting}[
    style=eprocode,
    label={lst:epro}
]
import torch

def epro_loss(log_u_a, log_p0_a, log_u_H, log_p0_H,
              log_q_H, advantage, mask, *, cap, tau, sigma):
    log_u_a, log_u_H = log_u_a.float(), log_u_H.float()

    with torch.no_grad():
        u = log_u_H.exp()
        p0 = log_p0_H.float().exp()
        q = log_q_H.float().exp()

        p_tail = (1 - p0.sum(-1)).clamp_min(1e-8)
        q_tail = (1 - q.sum(-1)).clamp_min(1e-8)
        u_tail = (1 - u.sum(-1)).clamp_min(1e-8)

        # 1. Execution-policy transport
        rho = q_tail / p_tail
        head = q * (u / p0.clamp_min(1e-8))
        Z = head.sum(-1) + rho * u_tail
        mu_hat_H = head / Z.unsqueeze(-1)

        center_coeff = (
            mu_hat_H - rho.unsqueeze(-1) * u / Z.unsqueeze(-1)
        )

        # 2. Policy Shift
        d = torch.cat([
            log_u_H - log_p0_H,
            (u_tail / p_tail).log().unsqueeze(-1)
        ], dim=-1)
        p = torch.cat([p0, p_tail.unsqueeze(-1)], dim=-1)
        p = p / p.sum(-1, keepdim=True)

        d_mean = (p * d).sum(-1, keepdim=True)
        shift = (p * (d - d_mean).square()).sum(-1).sqrt()

        # 3. Transported importance weight and SSC
        log_r = log_u_a.detach() - log_p0_a
        w = (log_r - Z.log()).exp().clamp(max=cap)
        outward = advantage * log_r > 0

        kappa = torch.where(
            outward,
            (-(shift - tau).clamp_min(0) / sigma).exp(),
            1.0
        )
        scale = advantage * w * kappa

    # 4. TSC surrogate loss
    correction = (center_coeff * log_u_H).sum(-1)
    loss = -scale * (log_u_a - correction)

    return (loss * mask).sum() / mask.sum().clamp_min(1)
\end{lstlisting}

\section{Execution-Policy Replay and Supplementary Results}\label{app:transport-measurement}
The experiment follows one recorded learner step, $\eta g_{\mathrm{RL}}=\theta_{\mathrm{end}}-\theta_0$, and evaluates $\theta(\alpha)=\theta_0+\alpha\eta g_{\mathrm{RL}}$. Thus $\alpha=1$ is the actual update, while larger values extrapolate along the same effective update direction (which can include the optimizer preconditioning). Prefixes and candidate-token coordinates remain fixed across evaluation scales. In the probability comparison, measured and transported execution probabilities are normalized on the same eight-token candidate sets. The six-panel results are shown in \cref{fig:execution_prediction_prob}.

The conditional-probability RMSE is $0.0087$ at $\alpha=1$, $0.0322$ at $\alpha=8$, and $0.0677$ at $\alpha=32$. The corresponding Pearson correlation is $0.9994$ at $\alpha=1$ and $0.9661$ at $\alpha=32$. These comparisons assess conditional next-token probabilities on a fixed candidate support; they should not be interpreted as full-vocabulary probability errors.

The centered-logit transport residual is
\begin{equation}\label{eq:residual}
\varepsilon_\theta(s)=\left[z_{\mu_\theta}(s)-z_{\mu_0}(s)\right]-\left[z_{\pi_\theta}(s)-z_{\pi_0}(s)\right],\qquad \varepsilon_{\theta_0}(s)=0.
\end{equation}
The residual summary in \cref{fig:execution_prediction_residual} uses ten measured update scales and 102,496 fixed token-coordinate trajectories from 12,812 held-out prefixes. Bands contain the central 50\%, 80\%, and 95\% of the measured residual distribution at each scale. A separate comparison with the unchanged execution anchor shows that the frozen anchor can have lower centered-logit point-prediction error at the immediate update endpoint, whereas at larger displacements the transported movement becomes more informative. For optimization, transport offers a moving, differentiable description of the execution policy rather than a static estimate.

\begin{figure}[t]
  \centering
  \includegraphics[width=\linewidth]{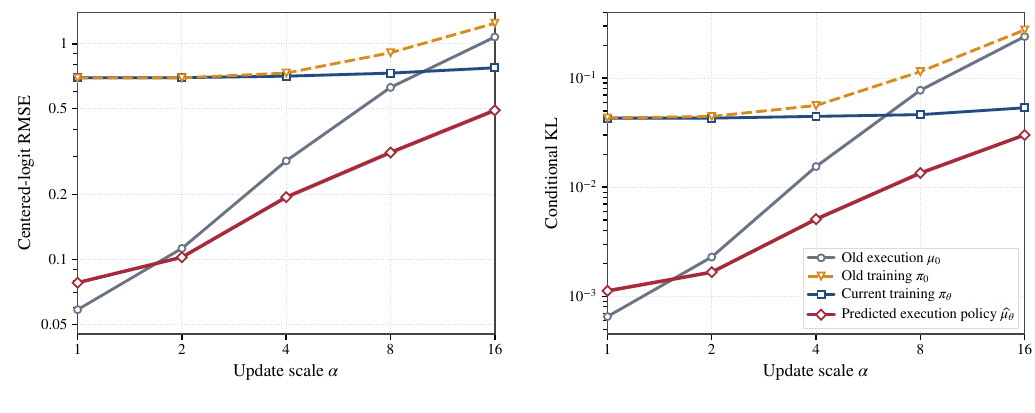}
  \caption{\textbf{Comparison of execution-policy predictors under simulated policy lag.}   For each update scale $\alpha$, we evaluate $\theta(\alpha)=\theta_0+\alpha\Delta\theta_{\mathrm{RL}}$   with the inference engine and measure how well each candidate matches the measured execution   policy $\mu_\theta$: (a) centered-logit RMSE; (b) conditional KL divergence   $D_{\mathrm{KL}}(\mu_\theta\,\|\,\cdot)$. The candidates are the   rollout-time execution policy $\mu_0$ (the centering distribution used by SC), the rollout-time   training policy $\pi_0$, the current training policy $\pi_\theta$, and the transported prediction   $\hat{\mu}_\theta$ from Eq.~(2).}
  \label{fig:predictor-comparison}
\end{figure}

\paragraph{Comparison with simpler predictors.}
Figure~\ref{fig:predictor-comparison} compares $\hat{\mu}_\theta$ with three simpler stand-ins for the current execution policy. The current training policy $\pi_\theta$ ignores the training--inference mismatch, so its error stays near the mismatch level for every $\alpha$. The rollout-time execution policy $\mu_0$ ignores policy lag, and its error grows roughly linearly in $\alpha$. The transported prediction combines both sources of information: its error grows sublinearly, and from $\alpha=2$ onward it is the most accurate of the four predictors. At $\alpha=16$, its centered-logit RMSE is about half that of $\mu_0$, and its conditional KL is about one eighth.

At $\alpha=1$, however, $\mu_0$ is more accurate than $\hat{\mu}_\theta$. We attribute this to the coarse resolution of the low-precision execution engine. A small parameter update leaves most quantized weights unchanged and moves a few of them by a full quantization step. The execution policy therefore responds sparsely and irregularly rather than following the smooth BF16 response that transport borrows. This is the regime in which the STE approximation is least accurate: quantization has zero derivative almost everywhere, and the quantized and BF16 responses agree only on average, once the accumulated displacement has moved many weights across quantization boundaries. The resulting gap is also small in absolute terms: about 0.02 in RMSE, compared with a mismatch level of roughly 0.7 for $\pi_\theta$.

\section{Prediction-Risk Evaluation Protocol}\label{app:risk-measurement}
For the risk-ranking experiments, measured transport error is the KL discrepancy between execution-engine probabilities and transported probabilities renormalized on their common candidate-token support. We restrict to held-out token positions with nonzero advantage and an outward update under the sampled-action ratio criterion. Within each update scale, the positive class is the highest 10\% of measured KL values. The continuous PPO ratio-based score and Policy Shift rank the same positions. AUC whiskers are 95\% paired prompt-bootstrap intervals over 15 informative prompt groups. AUC values in the main text correspond to $\alpha=1,2,4$ on this controlled replay, not to the end-to-end training runs.

\section{Baseline Update Formulations}\label{app:baselines}
We use GRPO and SC as shorthand for their token-level truncated importance-sampling (TIS) implementations, with importance-weight cap $C=2$.

\paragraph{GRPO.}
Our GRPO baseline uses group-normalized advantages in the decoupled PPO setting~\citep{hilton2022batchDecoupled,fu2026areal}:
\begin{equation}
\mathcal L_{\mathrm{PPO}}(\theta)
=-\mathbb E_t\left[
\bar\rho_t\min\left(
r_t(\theta)A_t,
\operatorname{clip}\bigl(r_t(\theta),1-\epsilon,1+\epsilon\bigr)A_t
\right)\right],
\end{equation}
where
\[
r_t(\theta)=\frac{\pi_\theta(a_t\mid s_t)}{\pi_0(a_t\mid s_t)},
\qquad
\bar\rho_t=\min\left(\frac{\pi_0(a_t\mid s_t)}{\mu_0(a_t\mid s_t)},C\right).
\]
The frozen training policy $\pi_0$ anchors PPO clipping, while $\bar\rho_t$ corrects for sampling from the execution policy $\mu_0$.
We set $\epsilon=0.2$ and additionally apply standard dual clipping with coefficient $3$ to negative-advantage terms.

\paragraph{SC and gated variants.}
SC uses conventional score centering with the truncated ratio $\pi_\theta/\mu_0$ and no PPO gate.
SC + PPO gate adds directional ratio gating: contributions are suppressed when $A_t>0$ and $r_t>1.2$, or $A_t<0$ and $r_t<0.8$.
TSC + PPO gate replaces the SC correction and importance weight with their transported counterparts, retaining the same gate.
EPRO retains TSC and replaces the PPO gate with Shift-Aware Soft Clipping (SSC).
The importance coefficients and SSC attenuation are held fixed during differentiation.

\section{Full Training Configuration}\label{app:training-config}
Each rollout iteration samples eight responses per prompt and is followed by one epoch of training in 16 learner minibatches.
We use temperature $1.0$, top-p $1.0$, no top-k sampling truncation, and maximum prompt/response lengths of 1,024/4,096 tokens.
All runs use BF16 training, group-normalized GRPO advantages, token-mean loss aggregation, gradient norm clipping at $1.0$, and a constant learning rate without warmup.
Entropy and KL regularization are disabled; AdamW uses betas $(0.9,0.999)$.
SC and transport-based estimators record a top-$32$ execution head. For Qwen3-1.7B-Base we use training data from DeepScaleR dataset~\citep{deepscaler2025}, and for the rest three models we use training data from GURU dataset~\citep{cheng2025revisitingGURU}. Main training hyperparameters are provided in \cref{tab:training-hyperparameters}.

\begin{table}[htbp]
\centering
\caption{Main training hyperparameters. Minibatch sizes are expressed in prompt groups; each group contains eight responses.}
\label{tab:training-hyperparameters}
\begingroup
\setlength{\tabcolsep}{4pt}
\renewcommand{\arraystretch}{1.12}
\small
\resizebox{\linewidth}{!}{%
\begin{tabular}{@{}lcccc@{}}
\toprule
& Qwen3-1.7B-Base & Qwen3-4B-Base & Qwen3-8B-Base & Qwen3-30B-A3B-Base \\
\midrule
Training data & DeepScaleR & GURU math & GURU math & GURU math \\
Optimizer & AdamW & AdamW & AdamW & SGD \\
Learning rate & $10^{-6}$ & $10^{-6}$ & $10^{-6}$ & $10^{-2}$ \\
Weight decay & $0.01$ & $0.01$ & $0.01$ & $0$ \\
Prompts per rollout & 256 & 256 & 128 & 128 \\
Minibatch size & 16 & 16 & 8 & 8 \\
Rollout weight precision & NVFP4 & NVFP4 & NVFP4 & FP8 \\
Rollout iterations & 450 & 400 & 300 & 300 \\
SSC threshold $\tau$ & $0.125$ & $0.125$ & $0.125$ & $0.3$ \\
SSC scale $\sigma$ & $0.0625$ & $0.0625$ & $0.0625$ & $0.15$ \\
\bottomrule
\end{tabular}%
}
\endgroup
\end{table}

The 30B-A3B setting additionally uses an NVFP4 KV cache.
For Qwen3-30B-A3B-Base, we deliberately disable Rollout Routing Replay (R3)~\citep{ma2025stabilizingR3} to study the high training--inference mismatch (TIM) regime.

\end{document}